\documentclass[10pt,twocolumn,table,letterpaper]{article}
\usepackage[pagenumbers]{iccv}

\usepackage{pifont}
\usepackage{makecell}
\usepackage{amsfonts, amssymb}

\definecolor{iccvblue}{rgb}{0.21,0.49,0.74}
\usepackage[pagebackref,breaklinks,colorlinks,allcolors=iccvblue]{hyperref}
\usepackage{float}

\title{Robust Fusion Controller: Degradation-aware Image Fusion with Fine-grained Language Instructions}

\author{Hao Zhang \footnotemark[1]\quad
	Yanping Zha \footnotemark[1]\quad
	Qingwei Zhuang\quad
	Zhenfeng Shao \quad
	Jiayi Ma \footnotemark[2] \\
	Wuhan University, Wuhan, China\\
	{\tt\small \{zhpersonalbox, jyma2010\}@gmail.com},\quad
	{\tt\small \{yanpingzha, zhuangqingwei, shaozhenfeng\}@whu.edu.cn}
}

\begin{document}
\maketitle
\footnotetext[1]{Equal Contribution}
\footnotetext[2]{Corresponding author}
\begin{abstract}
Current image fusion methods struggle to adapt to real-world environments encompassing diverse degradations with spatially varying characteristics. To address this challenge, we propose a robust fusion controller (RFC) capable of achieving degradation-aware image fusion through fine-grained language instructions, ensuring its reliable application in adverse environments. Specifically, RFC first parses language instructions to innovatively derive the functional condition and the spatial condition, where the former specifies the degradation type to remove, while the latter defines its spatial coverage. Then, a composite control priori is generated through a multi-condition coupling network, achieving a seamless transition from abstract language instructions to latent control variables. Subsequently, we design a hybrid attention-based fusion network to aggregate multi-modal information, in which the obtained composite control priori is deeply embedded to linearly modulate the intermediate fused features. To ensure the alignment between language instructions and control outcomes, we introduce a novel language-feature alignment loss, which constrains the consistency between feature-level gains and the composite control priori. Extensive experiments on publicly available datasets demonstrate that our RFC is robust against various composite degradations, particularly in highly challenging flare scenarios.
\end{abstract}
\begin{figure}[t]
	\centering
	\includegraphics[width=1\linewidth]{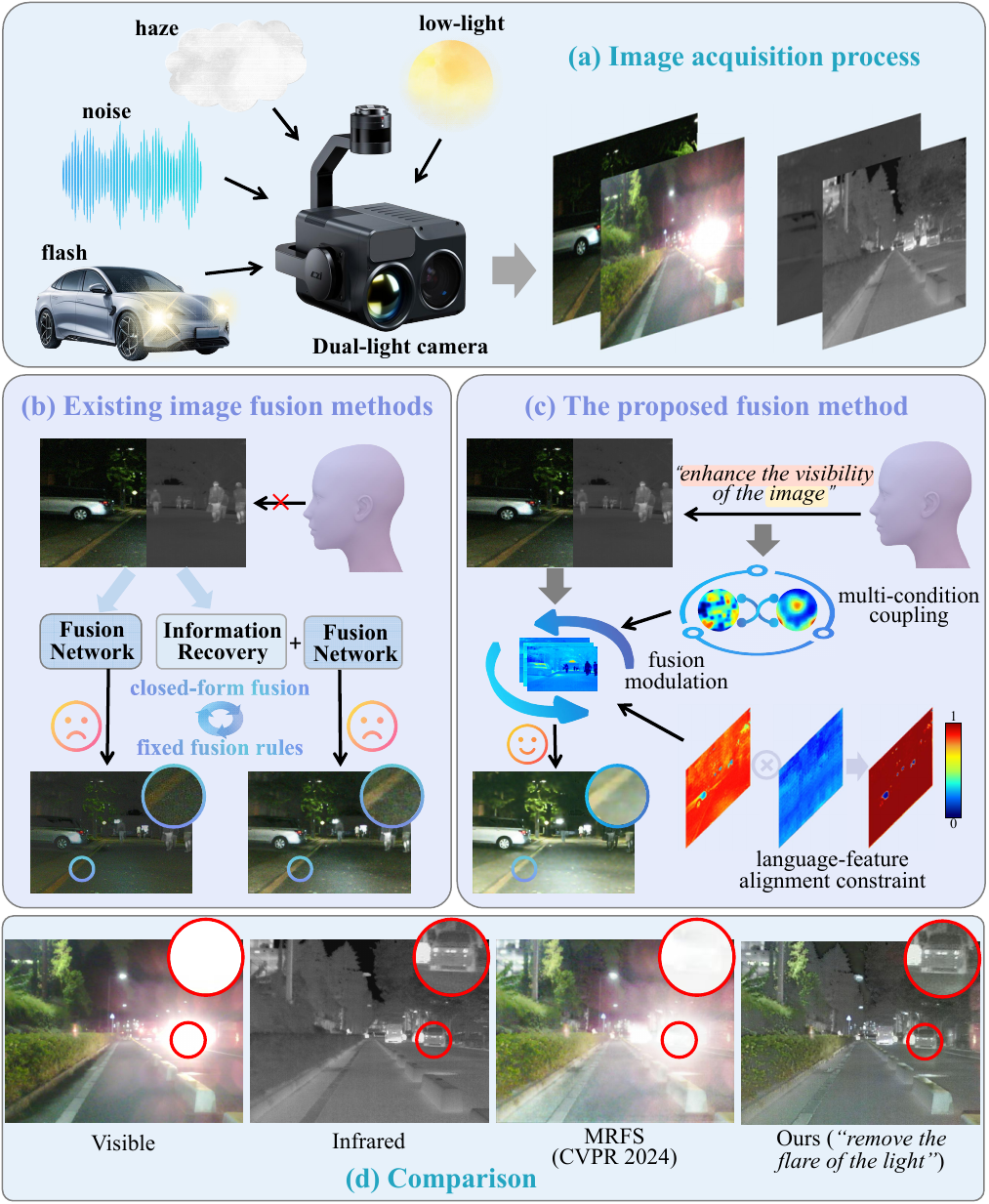} 
	\caption{Comparisons between our RFC and existing fusion methods in the degraded scenario.} 
	\label{fig:1}
\end{figure}
\section{Introduction}
\label{sec:intro}
Due to the limitation of the imaging principle, single-modal images can only capture partial scene attributes. For example, visible sensors can only capture texture and color information perceptible to human eyes, which cannot work in bad conditions, \textit{e.g.}, night, and haze. Infrared sensors can overcome adverse environments by capturing radiation information from thermal objects, but they often lack texture and color. In this context, image fusion technology emerges~\cite{singh2023review,huang2024leveraging,liu2024coconet}, aiming to integrate complementary information from multi-modal images to provide a comprehensive representation of the imaging scene. Thanks to this excellent representational capability, image fusion has become a core component of numerous intelligent perception applications, effectively enhancing the accuracy of military reconnaissance~\cite{muller2009cognitively}, autonomous driving~\cite{yadav2020cnn}, \textit{etc}.

From the task definition, the application scenarios of image fusion typically involve environments where a single sensor is ineffective due to poor conditions. In the real world, such environments typically exhibit two key characteristics. On the one hand, degradations are pervasive (\textit{e.g.}, overexposure, low-light, noise, haze, blur, and flare), with their types being diverse and severely compounded. On the other hand, these degradations exhibit spatially varying characteristics, potentially occurring both globally and locally. For instance, noise often appears in low-illumination regions, while flares typically accompany light sources. Therefore, equipping fusion models with the ability to overcome spatial-varying composite degradations is crucial for ensuring their reliable application in the real world.

Unfortunately, existing fusion methods struggle to meet this requirement, fundamentally hindering the practical application of image fusion technology. More concretely, mainstream fusion methods~\cite{li2018densefuse,zhang2023visible,ma2019fusiongan,tang2022image,liu2024task} that focus solely on enhancing information aggregation capabilities essentially do not eliminate degradations. Instead, the persistent presence of degradations leads to the erroneous discarding of valuable information, rendering image fusion more akin to a problem of ``\textit{information compression}". Differently, some of the latest methods~\cite{zhang2024dispel,chen2024lenfusion,zou2023infrared,zhang2024ev,tang2023divfusion} cooperate to achieve degradation removal and information fusion, enabling the restoration of more information from low-quality source images. This perspective tends to turn image fusion into a problem of ``\textit{information mining}". However, these methods can only handle a single type of degradation and are ineffective against composite degradations, let alone those with spatial variability.

To address these challenges, we propose a robust fusion controller, termed RFC. It derives a degradation-aware image fusion framework with fine-grained language instructions, enabling adaptability to harsh environments with spatial-varying composite degradations. Firstly, RFC parses language instructions to obtain two complementary control conditions. 1) \textbf{\textit{Functional condition}}: enables the specification of the degradation type to be removed, supporting both single-type degradation removal and unified removal of composite degradations. 2) \textbf{\textit{Spatial condition}}: defines the regions to be enhanced, supporting both local and global enhancement. Secondly, functional and spatial conditions are processed through a multi-condition coupling network, to generate composite control priori. This process translates abstract language instructions into latent control variables, providing a high-quality interactive medium for dynamically modulating the fusion process. Thirdly, the composite control priori is embedded into a hybrid attention-based fusion network through the linear feature modulation strategy~\cite{perez2018film}. While aggregating multi-modal information, it can precisely perceive and remove spatial-varying composite degradations. Finally, a novel language-feature alignment loss is introduced. By constraining the consistency between feature-level gains and the composite control priori, it can ensure that the controlled output aligns with the expectations of the language instructions. As presented in Fig.~\ref{fig:1}, our RFC significantly outperforms state-of-the-art methods in terms of harsh scenario characterization, particularly in challenging flare environments.

In summary, we make the following contributions:
\begin{itemize}
	\item We propose a robust fusion controller, forming a degradation-aware image fusion framework with fine-grained language instructions. To our knowledge, this is the first attempt in the field of image fusion to eliminate spatial-varying composite degradations, enhancing the robustness of fusion models in harsh environments.
	\item We design a novel generative mechanism for composite control priori, which can translate abstract language instructions into latent control variables. This enables us to establish an open-ended paradigm for image fusion, facilitating fine-grained functional control over arbitrary regions in accordance with user-defined instructions.
	\item A language-feature alignment loss is introduced, which drives feature gains to maintain potential consistency with composite control priori, strongly ensuring the modulation rationality of our RFC.
\end{itemize}

\begin{figure*}[t]
	\centering
	\includegraphics[width=0.95\linewidth]{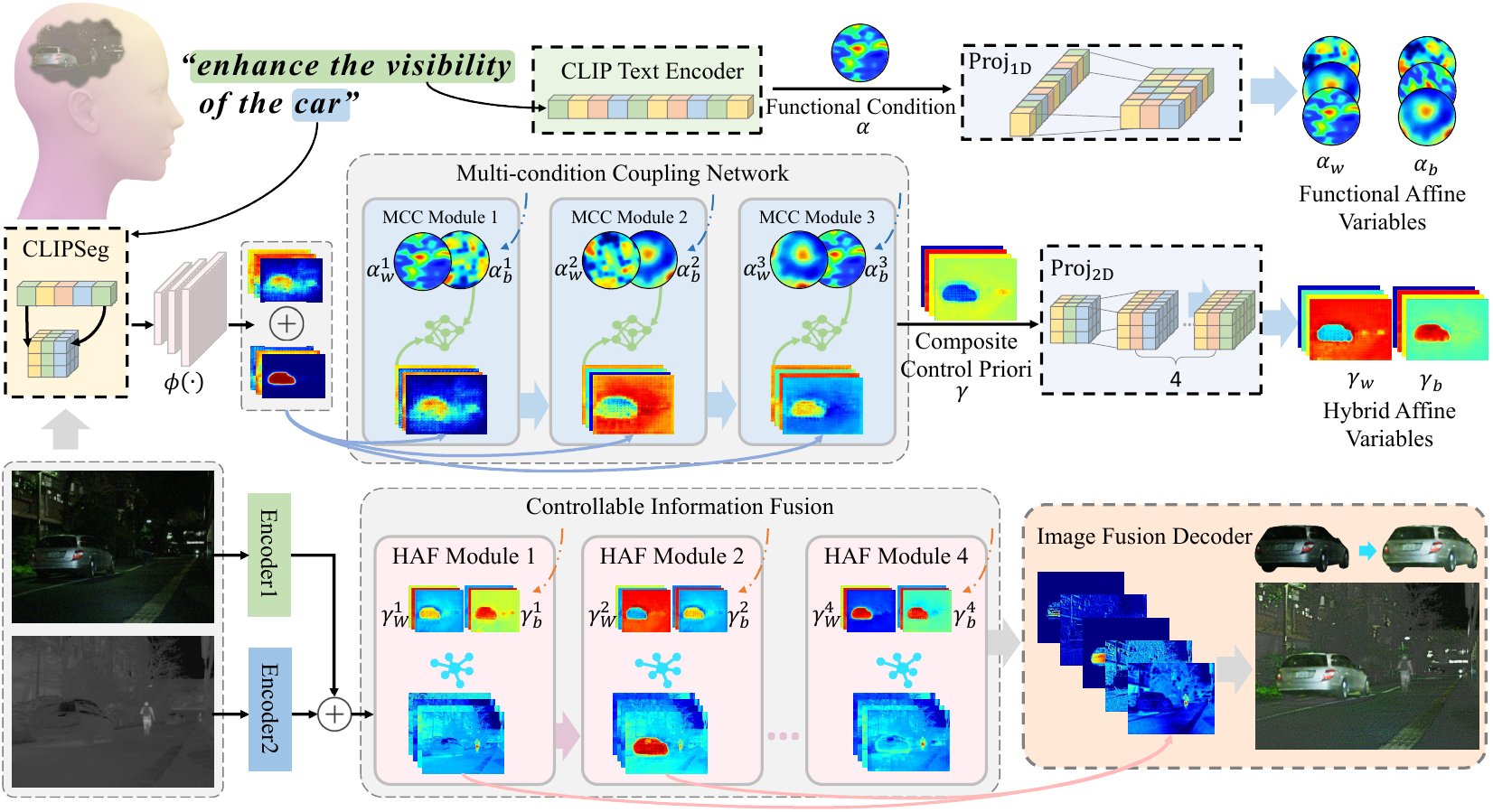} 
	\caption{The overall framework of our proposed RFC.} 
	\label{net}
\end{figure*}
\section{Related Work}
\label{sec:related work}
\subsection{Degradation-Robust Fusion}
Aware of the negative impact degradation has on image fusion, some of the latest methods have started focusing on improving the robustness of fusion models. For example, DDBF~\cite{zhang2024dispel} leverages infrared modality' high contrast to enhance low-light visible modality via cross-modal adversarial learning, while DIVFusion~\cite{tang2023divfusion} mitigates illumination degradation using the Retinex model in feature space. Both of them effectively enhance the robustness of fusion models in low-light environments. DeRUN~\cite{he2023degradation} integrates an optimization model with multi-stage processing to reduce noise while preserving textures. However, these methods are limited to addressing only a single type of degradation, which clearly falls short in real-world environments characterized by diverse degradations. Text-IF~\cite{yi2024text} uses text prompts to activate the fusion model for specific degradation types, but it is limited to addressing only one degradation in a single run, failing to handle scenarios with composite degradations. Thus, developing more robust fusion methods capable of handling composite degradations is highly desirable.

\subsection{Language-based Fusion}
The rapid development of vision-language large models (vLLM)~\cite{radford2021learning,li2022blip} has spurred methods integrating language into fusion processes to improve scene representation, categorized into language-assisted and language-controlled methods. Firstly, language-assisted fusion methods use scene-descriptive text as auxiliary information to enhance fusion representability. For example, FILM~\cite{zhao2024image} employs ChatGPT to generate descriptive text from source images, which is then encoded into the fusion process. Similarly, FTP~\cite{li2023text} leverages semantic distributions in text to enrich multi-modal feature interactions. However, these methods lack language-driven control, somewhat limiting interpretability. In contrast, language-controlled methods interact with the fusion model to dynamically adjust the process based on specific needs. For instance, the aforementioned Text-IF specifies degradation types via language input but only performs coarse global removal, failing to meet fine-grained requirements and handle spatial-varying composite degradations. By comparison, our method addresses users' diverse needs by generating composite control priori, enabling promising fusion control capabilities.

\section{Method}
Our RFC leverages language instructions to guide fusion, ensuring high-quality multi-modal aggregation while accurately removing spatial-varying composite degradations. We first parse instructions into functional and spatial conditions, defining the desired operation and target regions. These are then coupled into a composite control prior, modulating hybrid attention fusion modules to achieve the desired results. The overall framework is shown in Fig.~\ref{net}.

\subsection{Language Instruction Parsing}
Given the input language instruction $\zeta$, which expresses a composite requirement. First, we split $\zeta$ to obtain language fragments $\zeta_f$ that describe the functions (\textit{e.g.}, remove noise) and language fragments $\zeta_s$ that specify the spatial regions. The semantic content corresponding to them differs significantly, so we introduce two strategies to parse them separately. Specifically, $\zeta_f$ essentially represents a requirement for visual appearance, so we leverage the visual-text alignment capability of the CLIP~\cite{radford2021learning} model to parse it:
\begin{equation}
	     \alpha = E_{T}(\zeta_f),	
	\label{equ:1}
\end{equation}
where $E_{T}$ is the text encoder from the pre-trained CLIP, $\alpha \in \mathbb{R}^{B\times 512}$ indicates the obtained functional condition, and $B$ denotes the batch size. In contrast, $\zeta_s$ is more related to spatial localization, which cannot be handled by CLIP due to lacking fine-grained parsing capability. Thus, we introduce a powerful spatial parsing model, CLIPSeg~\cite{luddecke2022image}, for analyzing language fragments $\zeta_s$. Concretely, CLIPSeg is composed of transformer module $\Phi_t$ and convolution module $\Phi_c$ connected in series, which can locate specific regions in the images based on language instructions:
\begin{equation}
	\{S_{vis}, S_{ir}\} = \Phi_c(\Phi_t(\{I_{vis}, I_{ir}\}|\zeta_f)),
	\label{equ:2}
\end{equation}
where $\{I_{vis}, I_{ir}\}$ denotes the visible and infrared image pairs, and $\{S_{vis}, S_{ir}\}$ represent the spatial response maps output by CLIPSeg. However, CLIPSeg is trained only on the visible modality, which may reduce the location confidence on multi-modal data. Therefore, we fine-tune the convolutional module $\Phi_c$ to obtain a new one $\Phi^{'}_c$, and the fine-tuned spatial response maps are generated by:
\begin{equation}
	\{S^{'}_{vis}, S^{'}_{ir}\} = \Phi^{'}_c(\Phi_t(\{I_{vis}, I_{ir}\}|\zeta_f)).
	\label{equ:3}
\end{equation}
To account for the cross-modal priori knowledge before and after fine-tuning, we perform spatial response mixing to obtain a comprehensive spatial condition $\beta$:
\begin{equation}
	\beta = S_{vis}\oplus S_{ir}\oplus S^{'}_{vis}\oplus S^{'}_{ir},
	\label{equ:4}
\end{equation}
where $\oplus$ is the concatenation operation. Now, through parsing the input language instruction, we obtain the functional condition $\alpha$ and the spatial condition $\beta$.

\subsection{Composite Control priori Generation}
\begin{figure}[t]
	\centering
	\includegraphics[width=\columnwidth]{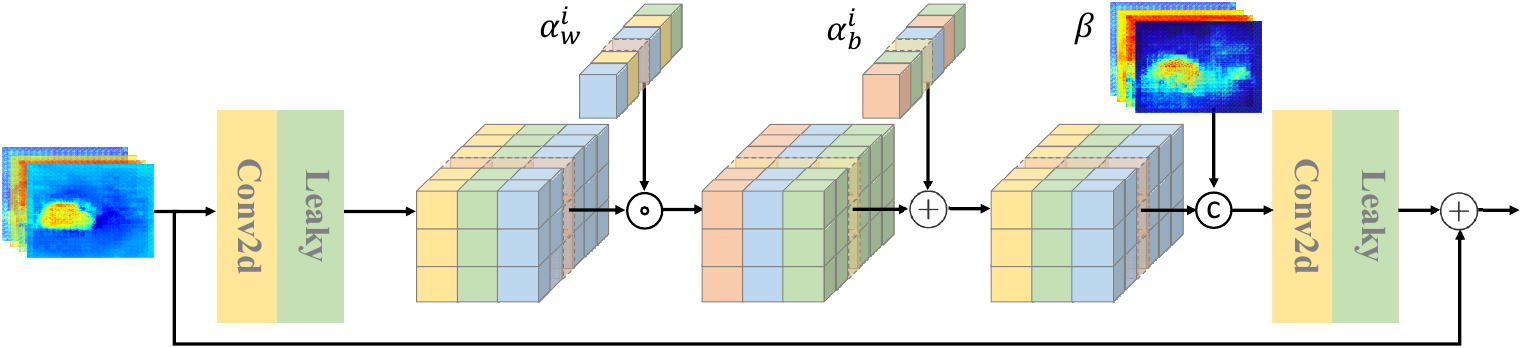}
	\caption{The architecture of multi-condition coupling module.}
	\label{resblock}
\end{figure}

Controlling the fusion process needs to be concentrated, implying the need for a control variable that can comprehensively represent both functional and spatial conditions. Inspired by FiLM~\cite{perez2018film}, we design multi-condition coupling (MCC) modules to combine $\alpha$ and $\beta$ following the idea of feature-wise affine transformation, as illustrated in Fig.~\ref{resblock}.

We first use two 1D convolution layers to generate functional affine variables from the functional condition $\alpha$: $\{\alpha_w^i, \alpha_b^i\} = \text{Proj}^{i}_{1D}(\alpha)$, where $\alpha_w^i$ indicates the functional weight, $\alpha_b^i$ denotes the functional bias, and $i$ is the index of the multi-condition coupling module. Then, we perform an affine transformation (AT) for a functional compound and combine it with the spatial condition: $\text{AT}(\cdot|\alpha_w^i, \alpha_b^i)\oplus \beta$. Such an operation ensures that the functional and spatial conditions are fully coupled, serving as the core component of the MCC module. The function of the MCC module can be represented as:
\begin{equation}
	F_{out}^i = \text{MCC}(F_{out}^{i-1},\text{AT}(\cdot|\alpha_w^i, \alpha_b^i)\oplus \beta),
\label{equ:5}
\end{equation}
where $F_{out}^{i-1}$ is the composite control prior output from the $i-1$-th MCC module, and when $i=1$, $F_{out}^{i-1}=\beta$. Totally, $3$ MCC modules are used, and the output $F_{out}^3$ of the last module is regarded as the composite control priori $\gamma$. It then undergoes 2D convolution to produce final hybrid affine variables: $\{\gamma_w^k, \gamma_b^k\} = \text{Proj}^{k}_{2D}(\gamma)$. At this point, $\{\gamma_w^k, \gamma_b^k\}$ can be considered to fully integrate both functional and spatial conditions.

  \begin{figure*}[t]
	\centering
	\includegraphics[width=0.99\textwidth]{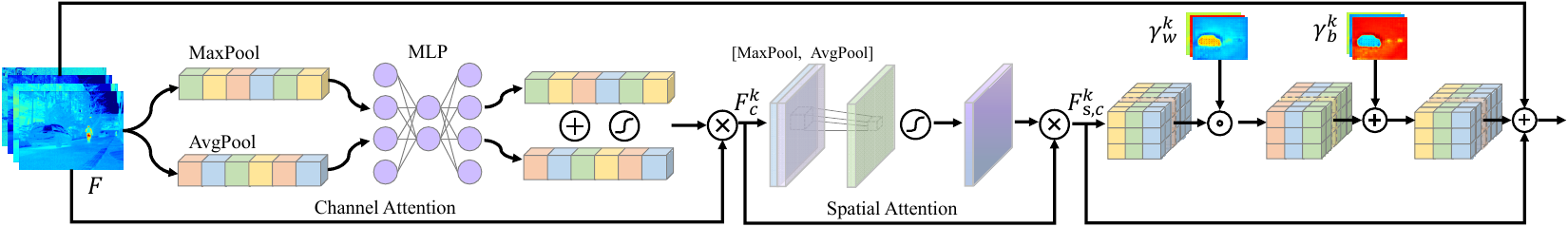}
	\vspace{-0.05in}
	\caption{The architecture of hybrid attention fusion module.}
	\label{cbablock}
\end{figure*}

\subsection{Controllable Information Fusion}
Next, the task at hand is to enable high-quality multi-modal feature fusion and seamlessly embed the generated composite control priori into the fusion process. We develop hybrid attention~\cite{woo2018cbam} fusion (HAF) modules to achieve this goal, as illustrated in \cref{cbablock}. First, we employ the channel attention mechanism to blend infrared and visible features. Formally, we use pooling operations to squeeze the input feature $F$, obtaining the maximum and average responses along the spatial dimensions, respectively. These responses are processed by a multi-layer perceptron (MLP), combined through summation, and subjected to a nonlinear activation to produce the final attention map, which is used to enhance the aggregated feature $F$. This process can be represented as:
\begin{equation}
	F_{c}^k = \sigma(\kappa(P_{M}(F^{k-1}))+\kappa(P_{A}(F^{k-1}))) \otimes F^{k-1},
	\label{equ:6}
\end{equation}
where $F^{k-1}$ is the feature output from the $k-1$-th HAF module, and when $k=1$, $F^{k-1}=F_{ir}\oplus F_{vis}$. $P_{M}$ and $P_{A}$ denote the maximum and average polling, $\kappa$ indicates the MLP function, and $\sigma$ indicates the Sigmoid function. Building on it, the spatial attention mechanism is utilized to reinforce the spatial representation of fused features $F_{c}^k$. Concretely, pooling operations are applied to $F_{c}^k$ to extract the maximum and average responses in the channel dimension, which are subsequently concatenated, projected, and activated to produce the spatial attention map. The spatial attention-based reinforcement process can be defined as:
\begin{equation}
	F_{c,s}^k = \sigma(\kappa(P_{A}(F_{c}^k)\oplus P_{M}(F_{c}^k))) \otimes F_{c}^k,
	\label{equ:7}
\end{equation}
where $F_{c,s}^k$ is the fused feature that has been attention-enhanced across both the spatial and channel dimensions. For embedding the composite control priori, we still follow the idea of feature-wise affine transformation. Specifically, we use the affine variables $\{\gamma_w^k, \gamma_b^k\}$ to process $F_{c,s}^k$:
\begin{equation}
	F_{control}^k = \text{AT}(F_{c,s}^k|\gamma_w^k, \gamma_b^k)+F^{k-1},
	\label{equ:8}
\end{equation}
Finally, an UNet-like~\cite{ronneberger2015u} decoder $D_U$ with skip connections is employed to reconstruct the fused image: $	I_{f}=D_U(F_{control}^1, F_{control}^2, F_{control}^3, F_{control}^4)$.

\subsection{Optimization Regularization}
The above designs offer the architectural support for robust image fusion with fine-grained language instructions. To ensure their effective operation, we formulate optimization regularization, comprising a degradation-aware reconstruction loss and a language-feature alignment loss.

\noindent\textbf{Degradation-aware Reconstruction Loss}. This regularization term aims to drive the targeted removal of degradations, enhancing perceptual fidelity. The data used to construct this loss is multi-modal clean-degraded image pairs $\{I_{vis},I_{ir}, I^{'}_{vis},I^{'}_{ir}\}$, where $I_{vis}$ and $I_{ir}$ are degraded images, and $I^{'}_{vis}$ and $I^{'}_{ir}$ are corresponding clean ones.

Based on the input language instruction $\zeta$, we identify two conditions: the degradation type $\Omega$ (\textit{e.g}, low light, overexposure (flare), haze, noise, blur, and their composites) and the target region $\Lambda$ (can be either a local region or the entire image). Firstly, According to $\Omega$, we retrieve  $\{I^{\Omega}_{vis},I^{\Omega}_{ir}\}$ that includes this specific degradation from the dataset. Notably, $\Omega$ can represent a compound of multiple types of degradation. Secondly, in conjunction with $\Lambda$, we simulate pseudo multi-modal references:
\begin{equation}
    \{\hat{I}_{vis},\hat{I}_{ir}\} = \{I^{\Omega}_{vis},I^{\Omega}_{ir}\}_{\overline{\Lambda}}+\{I^{'}_{vis},I^{'}_{ir}\}_{\Lambda} ,
	\label{equ:9}
\end{equation}
where $\overline{\Lambda}$ indicate regions that are not specified by language instruction $\zeta$. With the pseudo multi-modal references in place, we constrain the final fused image $I_{f}$ from three aspects: contrast, structure, and color. The corresponding loss functions are defined as:
\begin{equation}
	\mathcal{L}_{con} \!\!=\!\!\! \sum\nolimits\!\alpha_{\{\Lambda, \overline{\Lambda}\}}\lVert I^{y}_{f}\!-\!\text{max}(\!\{\hat{I}^{y}_{vis},\!\hat{I}_{ir}\}\!)\rVert_{\{\Lambda, \overline{\Lambda}\}},  \label{equ:10}
\end{equation}
\vspace{-0.2in}
\begin{equation}
	\mathcal{L}_{str} \!\!=\!\! \sum\nolimits\!\alpha_{\{\Lambda, \overline{\Lambda}\}}\lVert \nabla I^{y}_{f}\!-\!\text{max}(\! \{\nabla\hat{I}^{y}_{vis},\!\nabla\hat{I}_{ir}\}\!)\rVert_{\{\Lambda, \overline{\Lambda}\}},  \label{equ:11}
\end{equation}
\vspace{-0.1in}
\begin{equation}
	\mathcal{L}_{cor} = \sum\nolimits\!\alpha_{\{\Lambda, \overline{\Lambda}\}} \lVert I^{cbcr}_{f}-\hat{I}^{cbcr}_{vis}\rVert_{\{\Lambda, \overline{\Lambda}\}},  \label{equ:12}
\end{equation}
where superscripts $y$ and $cbcr$ denote the illumination and chrominance channels, respectively. We use dynamic weights $\alpha_{\{\Lambda, \overline{\Lambda}\}}$ to distinctively handle the distance calculations for the language-specified region $\Lambda$ and other regions $\overline{\Lambda}$. The dynamic weights are defined as $\alpha_\Lambda = \Upsilon(I_f)/\Upsilon(\Lambda)$, in which $\Upsilon$ is an operator that calculates the number of pixels in specific regions. Such a mechanism can effectively prevent the limitation of the target region from being overlooked during the optimization process when its size is too small.

\begin{figure}[t]
	\centering
	\includegraphics[width=0.98\columnwidth]{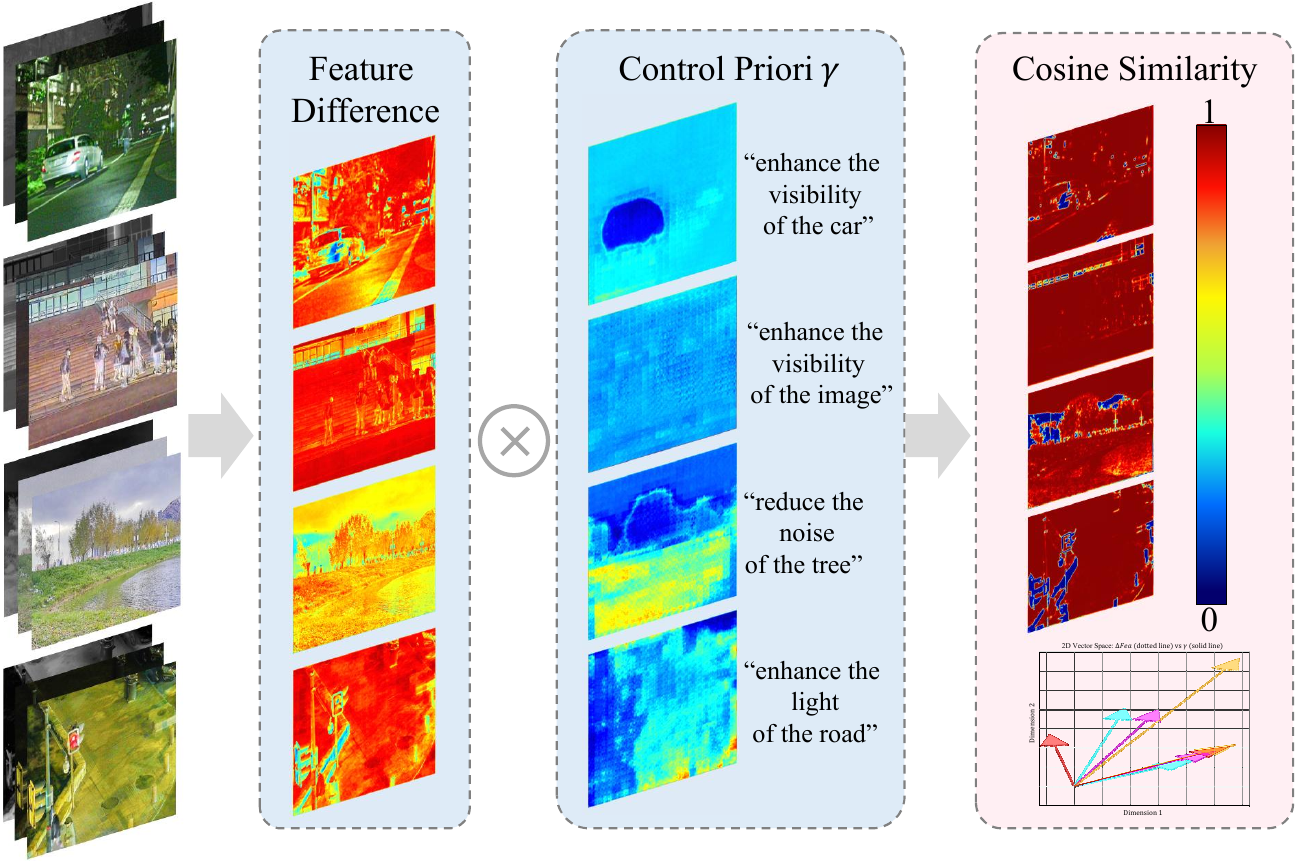}
	\vspace{-0.08in}
	\caption{Schematic diagram of the alignment mechanism between feature-level gains and language instructions.}
	\label{loss}
\end{figure}

\begin{figure*}[t]
	\centering
	\includegraphics[width=1\textwidth]{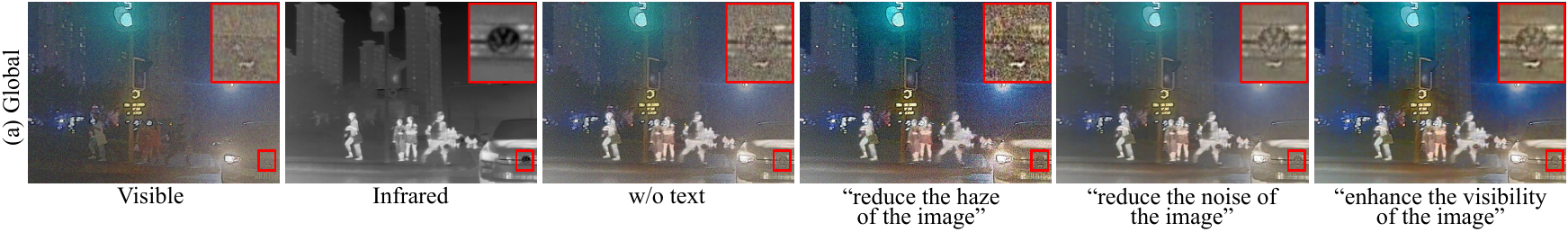}
	\includegraphics[width=1\textwidth]{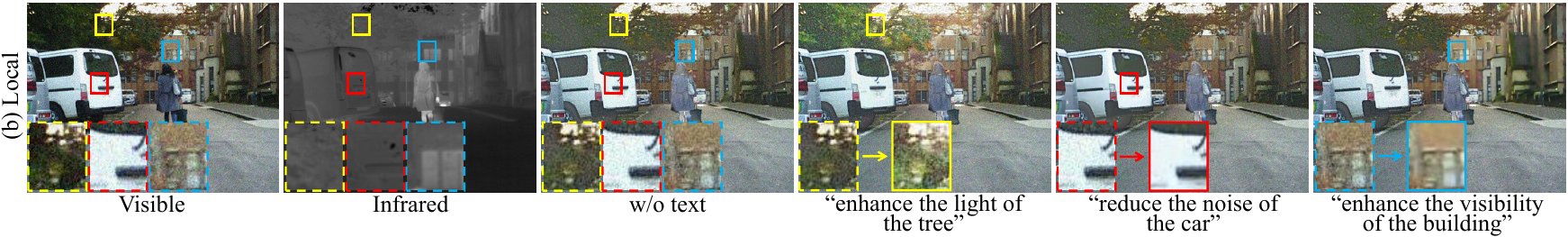}
	\vspace{-0.8cm}
	\caption{Demonstrating global and local degradation-aware fusion ability of our RFC.}
	\label{gl-con}
\end{figure*}

\begin{figure}[t]
	\centering
	\includegraphics[width=1\columnwidth]{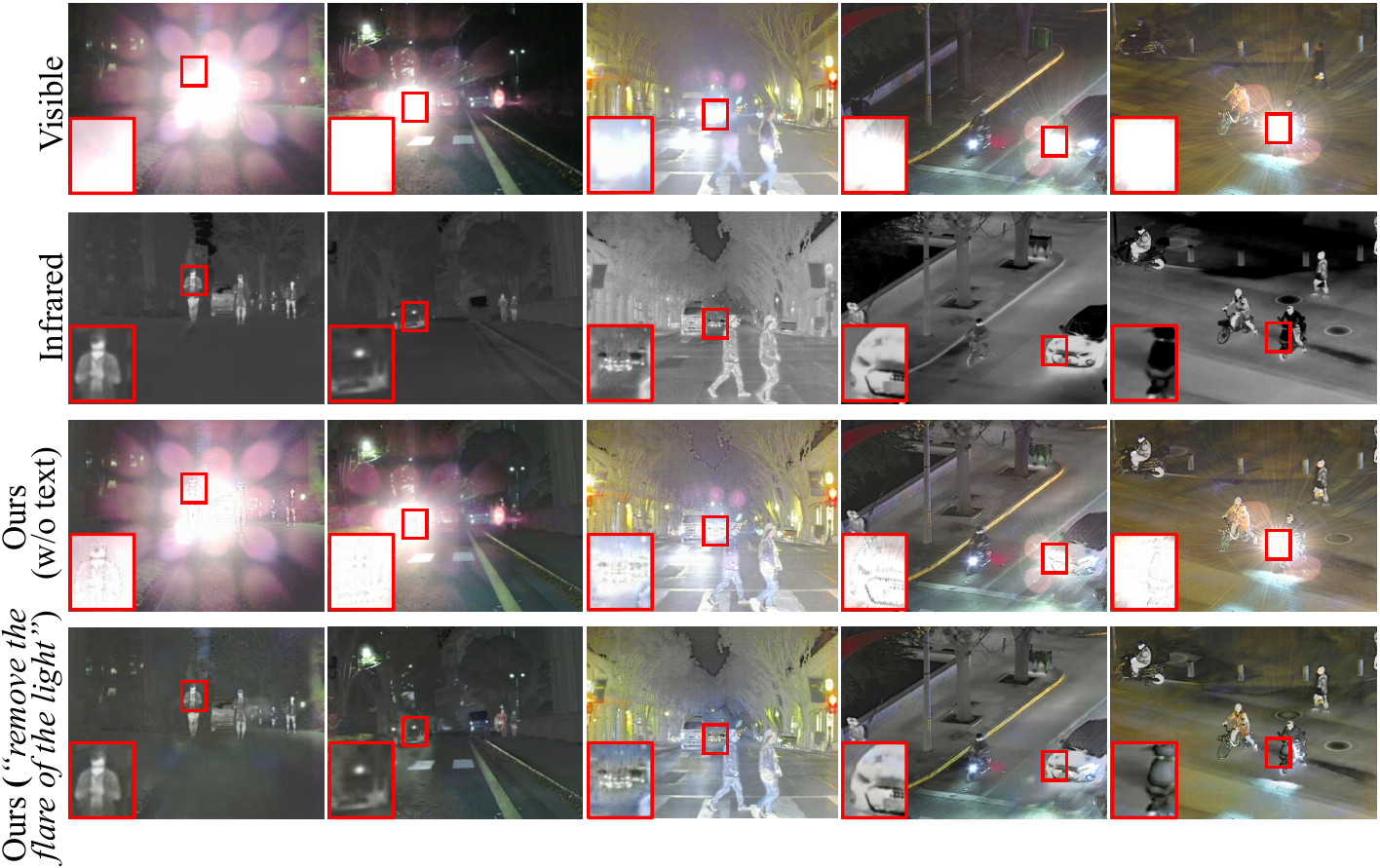}
	\vspace{-0.8cm}
	\caption{The flare removal function of our RFC.}
	\label{flare}
\end{figure}

\noindent\textbf{Language-Feature Alignment Loss}. Through the above loss, the dynamic responsiveness of the final fused result to language instructions can be effectively driven. However, the internal fusion process still lacks constraints, which potentially compromises the fusion model's sensitivity to language instructions. To this end, we introduce a language-feature alignment loss, primarily ensuring that the feature-level gains introduced by HAF modules remain consistent with the composite control priori. As demonstrated in Fig.~\ref{loss}, we calculate the difference between the input of the first HAF module and the output of the final HAF module, representing the feature-level gains modulated by the composite control priori: $\Delta{F} = F_{control}^0-F_{control}^4=F_{ir}\oplus F_{vis}-F_{control}^4$. Then, the language-feature alignment loss is defined as:
\begin{equation}
	\mathcal{L}_{ali} = 1 - \frac{\langle \tau(\gamma), \tau(\Delta F) \rangle}{|\tau(\gamma)| \times |\tau(\Delta F|)}, \label{equ:13}
\end{equation}
where $\tau$ is the flattening operator, and $\langle \cdot \rangle$ is the vector dot product. This loss effectively ensures the rationality of the intermediate fusion processes under language instructions.

\section{Experiments}
\subsection{Experimental Configurations}
\noindent\textbf{Datasets}. We construct the required training and testing dataset based on MFNet~\cite{ha2017mfnet}, LLVIP~\cite{jia2021llvip}, M3FD~\cite{liu2022target}, FMB~\cite{liu2023multi}, and RoadScene~\cite{xu2020fusiondn} datasets. Specifically, we extend these existing datasets with simulated degradations (\textit{e.g}, low light, overexposure (flare), haze, noise, blur, and their composites). Our training set includes $14,654$ image-text pairs with annotations specifying degradation types and regions. Testing employs $700$ multi-modal image pairs.

\noindent \textbf{Implementation}. We use the AdamW optimizer with an initial learning rate $2e^{-4}$ to update the parameters of all network modules. All experiments are conducted on four NVIDIA Tesla P100 GPUs with $16$ GB memory and one Intel(R) Xeon(R) Gold 5117 CPU.

\begin{figure*}[t]
	\centering
	\includegraphics[width=\textwidth]{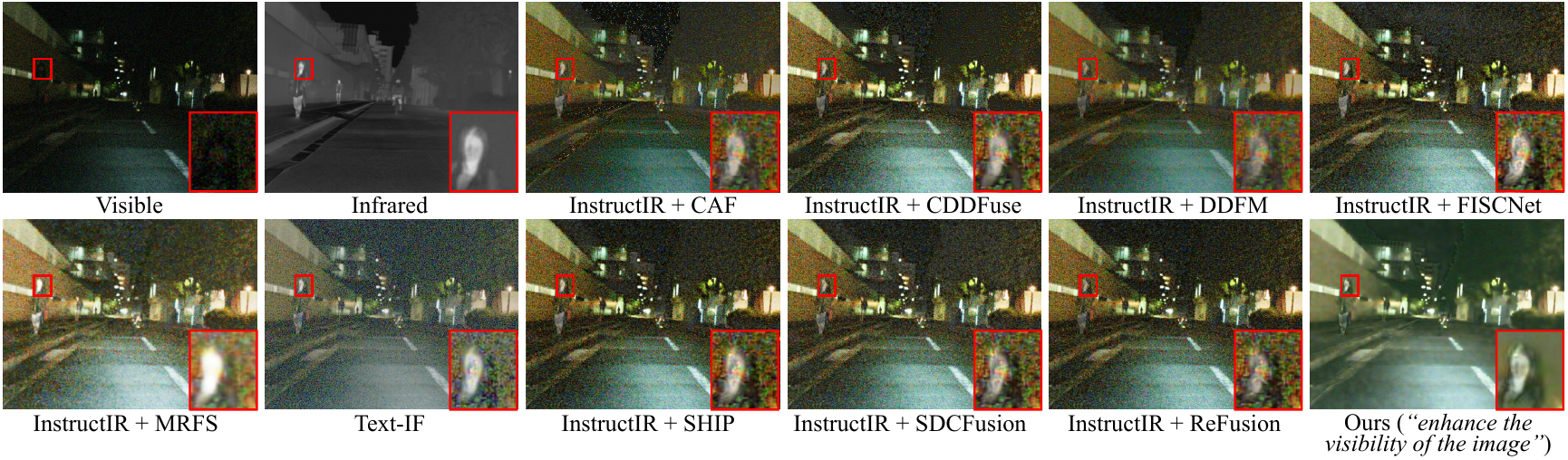}
	\vspace{-0.8cm}
	\caption{Qualitative results in scenarios with composite degradations. For more results, please see the supplementary material.}
	\label{MIX}
\end{figure*}

\begin{table*}[ht]
	\caption{Quantitative results in scenarios with composite degradations. (\textbf{Bold} is the best, \underline{underlined} is the second). For more results, please see the supplementary material.}
	\vspace{-0.3cm} \label{MIXtable2}
	\resizebox{\textwidth}{!}
	{
		\begin{tabular}{lcccccc|lcccccc}
			\Xhline{1.3pt}
			\rule{0pt}{2.8ex}
			\textbf{M3FD} & Qabf                         & SSIM                         & SD                            & MI                           & VIF                          & SCD                          & \textbf{MFNet}     & Qabf                         & SSIM                         & SD                            & MI                           & VIF                          & SCD                          \\ \Xhline{1.15pt} \rule{0pt}{2.8ex}
			DDFM                              & 0.244                                            & 0.283                                            & 31.421                                            & 1.861                                            & 0.326                                            & \cellcolor{pink!40} \textbf{ 1.562} & DDFM                      & 0.284                        & 0.237                        & 34.463                        & 2.093                        & 0.526                        & \cellcolor{blue!10} \underline{1.392} \\
			SHIP                              & 0.417                                            & 0.285                                            & 35.353                                            & 2.052                                            & 0.298                                            & 1.081                        & SHIP                      & 0.340                        & 0.239                        & 41.800                        & 2.230                        & 0.540                        & 1.088                        \\
			CDDFuse                           & 0.392                                            & 0.278                                            & { 38.967}                     & 1.983                                            & 0.307                                            & { 1.398} & CDDFuse                   & 0.340                        & 0.254                        & \cellcolor{pink!40} \textbf{ 46.574} & { 2.488} & 0.538                        & 1.175                        \\
			MRFS                              & 0.039                                            & 0.181                                            & \cellcolor{blue!10} \underline{39.591}                     & \cellcolor{blue!10} \underline{2.208}                     & { 0.328}                     & 1.139                        & MRFS                      & 0.299                        & { 0.264} & { 45.194} & \cellcolor{blue!10} \underline{2.560} & 0.478                        & 1.217                        \\
			CAF                               & 0.307                                            & { 0.289}                     & 33.321                                            & 1.872                                            & 0.310                                            & \cellcolor{blue!10} \underline{1.517} & CAF                       & 0.292                        & \cellcolor{blue!10} \underline{0.268} & 29.288                        & 1.805                        & { 0.554} & { 1.271} \\
			ReFusion                          & {0.437}                        & \cellcolor{pink!40} \textbf{ 0.290} & {37.148}                        &  {1.981}                        &  {0.322}                        &  {1.331}    & ReFusion                  & 0.347                        & 0.247                        & 45.102                        & 2.287                        & 0.547                        & 1.212                        \\
			FISCNet                           &  {{ 0.442}} &  {0.276}                        &  {36.692}                        &  {2.011}                        &  {0.302}                        &  {1.073}    & FISCNet                   & { 0.352} & 0.234                        & 43.889                        & 2.171                        & 0.521                        & 1.051                        \\
			SDCFusion                         &  {\cellcolor{blue!10} \underline{0.446}} &  {0.286}                        &  {34.914}                        &  {1.898}                        &  {0.306}                        &  {1.274}    & SDCFusion                 & \cellcolor{blue!10} \underline{0.366} & 0.258                        & 43.400                        & 2.194                        & \cellcolor{pink!40} \textbf{ 0.596} & 1.237                        \\
			Text-IF                           & {0.396}                        & {0.234}                        & {36.757}                        & {{ 2.118}} & {\cellcolor{blue!10} \underline{0.329}} &  {1.087}    & Text-IF                   & 0.291                        & 0.154                        & 39.336                        & 2.002                        & 0.304                        & 1.070                        \\
			Ours                              &  {\cellcolor{pink!40} \textbf{ 0.452}} &  {\cellcolor{blue!10} \underline{0.289}} &  {\cellcolor{pink!40} \textbf{ 43.746}} &  {\cellcolor{pink!40} \textbf{ 2.528}} &  {\cellcolor{pink!40} \textbf{ 0.378}} &  {1.095}    & Ours                      & \cellcolor{pink!40} \textbf{ 0.400} & \cellcolor{pink!40} \textbf{ 0.316} & \cellcolor{blue!10} \underline{45.674} & \cellcolor{pink!40} \textbf{ 2.692} & \cellcolor{blue!10} \underline{0.582} & \cellcolor{pink!40} \textbf{ 1.396} \\ \Xhline{1.3pt}
		\end{tabular}
	}
\end{table*}

\begin{figure*}[!ht]
	\centering
	\includegraphics[width=\textwidth]{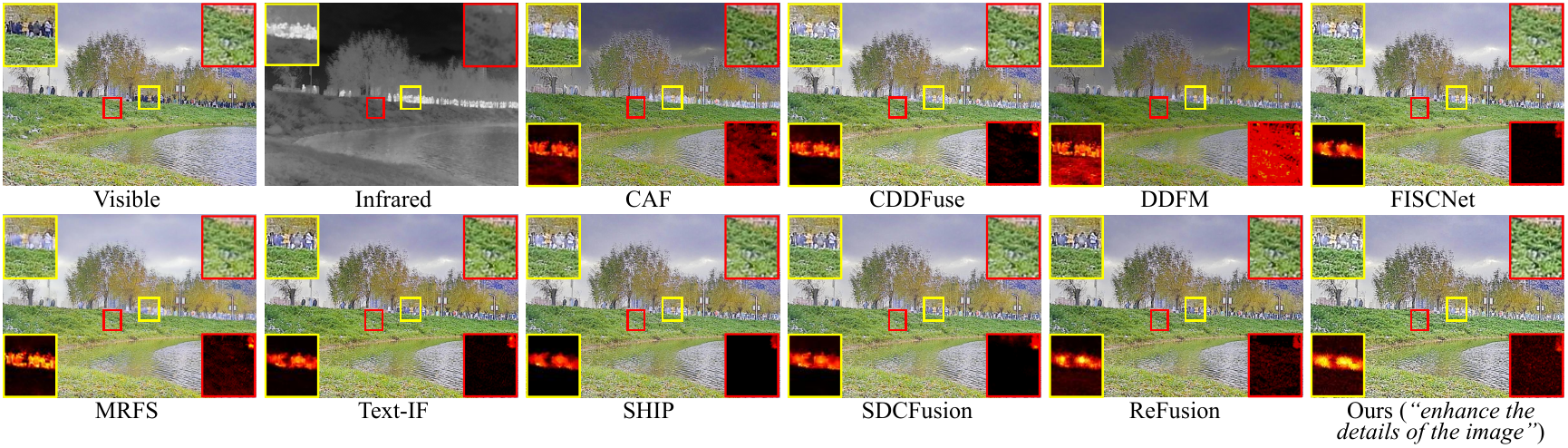}
	\vspace{-0.8cm}
	\caption{Qualitative results in scenarios without degradations. The highlighted area at the bottom represents the residual maps between fused results and the source visible image.}
	\label{Defig}
\end{figure*}

\subsection{Functional Validation}
First, we demonstrate the powerful fine-grained degradation removal capability of our RFC during the fusion process, which operates both globally and locally.

\noindent\textbf{Global Degradation-aware Fusion}. As shown in Fig.~\ref{gl-con} (a), our RFC achieves global degradation removal under different language instructions. For instance, instructions like ``reduce the noise" selectively suppress noise while preserving other features, whereas ``enhance the visibility" eliminates composite degradations, producing a completely clean output. This highlights RFC’s capability to interpret language instructions for targeted enhancement precisely.

\noindent\textbf{Local Degradation-aware Fusion}. Our RFC supports local degradation removal, addressing practical needs such as enhancing key objects (\textit{e.g.}, cars, pedestrians). As shown in Fig. \ref{gl-con} (b), when an instruction specifies both the degradation type and target region, RFC selectively restores the designated areas while maintaining contextual consistency.

\noindent\textbf{Flare Removal}. In night-time driving scenarios, lens flare degrades image quality, impairing visibility. As a highlight, our RFC mitigates flare artifacts through language-driven modulation in Fig.~\ref{flare}, leveraging infrared cues to compensate for overexposed regions. This results in robustness improvements to produce perceptually superior fused results.

\subsection{Comparison with Composite Degradations}
We compared RFC with nine SOTA methods: MRFS~\cite{zhang2024mrfs}, CDDFuse~\cite{zhao2023cddfuse}, DDFM~\cite{zhao2023ddfm}, CAF~\cite{liu2024elegance}, SHIP~\cite{zheng2024probing}, FISCNet~\cite{zheng2024frequency}, ReFusion~\cite{bai2024refusion}, SDCFusion~\cite{liu2024semantic}, and Text-IF~\cite{yi2024text}. For methods without degradation removal capabilities, we use an all-in-one enhancement method, InstructIR~\cite{conde2024instructir} for pre-processing. For Text-IF, we inform it of all the types of degradation present in the source images through textual input. Our RFC is tested with the default instruction: ``enhance the visibility of the image". Visual results in Fig.~\ref{MIX} highlight RFC's advantages in handling composite degradations. For example, the competitors all fail to handle the noise that is introduced when enhancing illumination. In contrast, our RFC effectively removes composite degradations, while preserving the saliency of the pedestrian and the fine details in the background. Furthermore, the quantitative results in Table~\ref{MIXtable2} show that our RFC outperforms other methods on most metrics, demonstrating its ability to retain key scene information.

\subsection{Comparison without Degradations}
Beyond removing degradations, our RFC can further strengthen scene details in degradation-free environments. Our RFC is tested under the instruction ``enhance the details of the image". As shown in Fig.~\ref{Defig}, the residual maps between fused results and the source visible image reveal that RFC retains more high-frequency information in areas like trees and grass, while preserving richer thermal radiation in the human region. Since RFC achieves information generation beyond source images in degradation-free scenarios, full-reference metrics are no longer used. Instead, we employ four no-reference metrics to evaluate qualitative performance, as shown in Table~\ref{Detable}. It can be observed that our method still achieves the best scores on most metrics.

\subsection{Generalization Experiment}
\begin{figure}[t]
	\centering
	\includegraphics[width=\columnwidth]{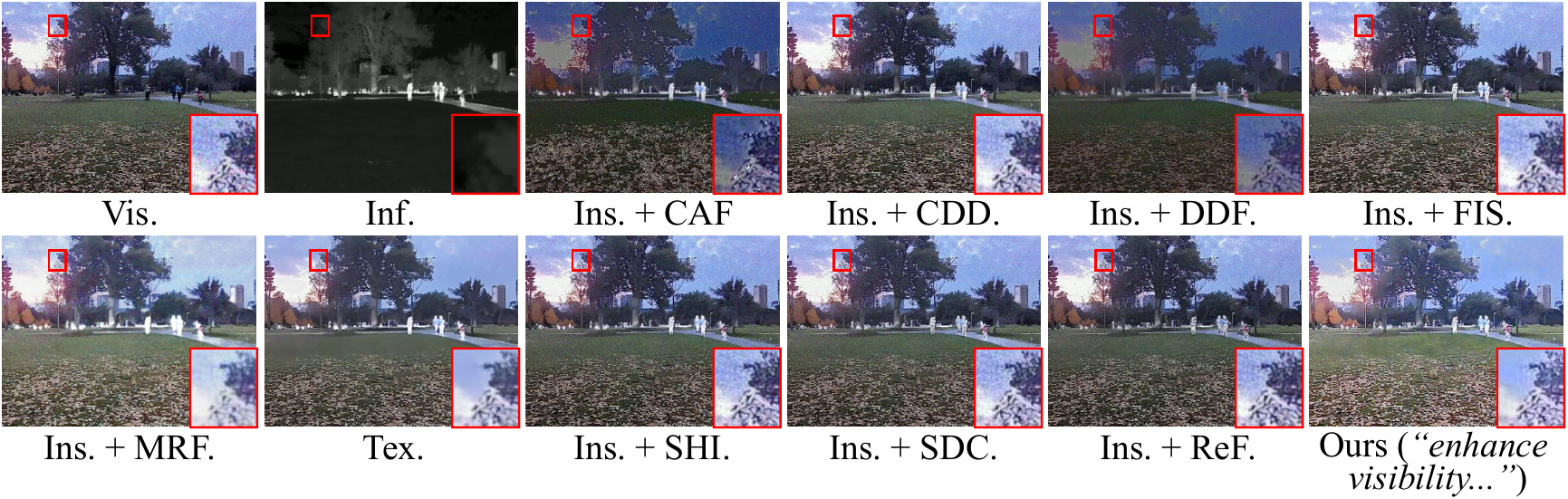}
	\vspace{-0.8cm}
	\caption{Qualitative results of generalization experiment.}
	\label{gene}
\end{figure}

Furthermore, we conduct generalization experiments on the M2VD~\cite{tang2025controlfusioncontrollableimagefusion} dataset. As shown in Fig.~\ref{gene} and Table~\ref{m3svd-tab}, our RFC outperforms other methods in both visual quality and objective scores, demonstrating its strong generalization ability and effectiveness across diverse fusion scenarios.

\begin{table}[t]
    \renewcommand \arraystretch{1.1}
	\caption{Quantitative results in scenarios without degradations.}
	\vspace{-0.3cm} \label{Detable}
	\resizebox{\columnwidth}{!}
	{
		\begin{tabular}{c|cccccccccc}
			\Xhline{1.3pt}
			\rule{0pt}{2.8ex}
			\textbf{M3FD} & DDF.    & SHI.    & CDD.   & MRF. & CAF    & ReF.  & FIS.                        & SDC. & Tex.                       & Ours                          \\ 		\Xhline{1.15pt}
			\rule{0pt}{2.8ex}
			SD                                & 33.15 & 43.42 & { 45.20} & 43.22 & 38.23 & 44.61   & 45.17                        & 42.67    & \cellcolor{blue!10} \underline{46.70} & \cellcolor{pink!40} \textbf{ 47.44} \\
			AG                                & 5.45  & 8.99  & 8.80                         & 6.66  & 7.64  & 9.21    & \cellcolor{blue!10} \underline{9.51}  & 8.99     & { 9.36}  & \cellcolor{pink!40} \textbf{ 12.21} \\
			EN                                & 6.89  & 7.22  & { 7.27}  & 7.17  & 7.11  & 7.27    & 7.27                         & 7.20     & \cellcolor{pink!40} \textbf{ 7.33}  & \cellcolor{blue!10} \underline{7.33}  \\
			SF                                & 14.07 & 23.71 & 23.60                        & 18.24 & 21.49 & 24.34   & \cellcolor{blue!10} \underline{25.00} & 23.64    & { 24.78} & \cellcolor{pink!40} \textbf{ 31.08} \\ \Xhline{1.3pt}
		\end{tabular}
	}
\end{table}

\begin{table}[t]
    \renewcommand \arraystretch{1.1}
	\caption{Quantitative results of generalization experiment.}
	\vspace{-0.3cm}  \label{m3svd-tab}
	\resizebox{\columnwidth}{!}
	{
		\begin{tabular}{c|cccccccccc}
			\Xhline{1.3pt}
			\rule{0pt}{2.8ex}
			\textbf{M2VD} & DDF.    & SHI.    & CDD.   & MRF. & CAF    & ReF.  & FIS.                        & SDC. & Tex.                       & Ours      \\ 	\Xhline{1.15pt}
			\rule{0pt}{2.8ex}
			Qabf           & 0.30                        & 0.48  & 0.47                         & 0.42                        & 0.24  & 0.48                        & 0.49                         & {0.49} & \cellcolor{blue!10} \underline{0.52}  & \cellcolor{pink!40} \textbf{ 0.54} \\
			SSMI           & 0.26                        & 0.22  & 0.22                         & \cellcolor{blue!10} \underline{0.30} & 0.19  & 0.24                        & 0.23                         & {0.26} & 0.26                         & \cellcolor{pink!40} \textbf{ 0.31} \\
			SD             & 38.25                       & 64.70 & \cellcolor{pink!40} \textbf{ 67.97} & 62.96                       & 46.79 & 66.43                       & \cellcolor{blue!10} \underline{67.51} & 65.68                       & {66.86} & 59.75                       \\
			MI             & 2.39                        & 2.60  & \cellcolor{blue!10} \underline{2.77}  & \cellcolor{pink!40} \textbf{ 2.98} & 1.97  & {2.76} & 2.68                         & 2.62                        & 2.68                         & 2.44                        \\
			VIF            & 0.35                        & 0.33  & 0.33                         & \cellcolor{blue!10} \underline{0.39} & 0.21  & 0.34                        & 0.34                         & 0.35                        & {0.38}  & \cellcolor{pink!40} \textbf{ 0.40} \\
			SCD            & {1.34} & 1.09  & 1.13                         & \cellcolor{pink!40} \textbf{ 1.45} & 1.02  & 1.14                        & 1.12                         & 1.25                        & 1.19                         & \cellcolor{blue!10} \underline{1.39} \\ \Xhline{1.3pt}
		\end{tabular}
	}
\end{table}

\begin{figure}[t]
	\centering
	\includegraphics[width=\columnwidth]{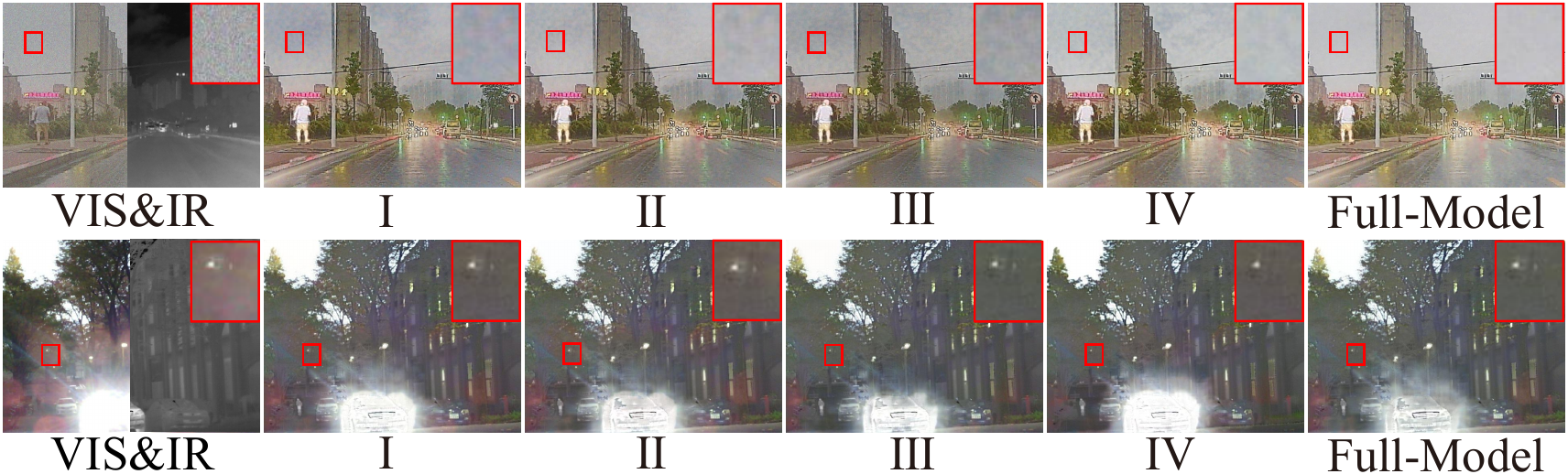}
	\vspace{-0.7cm}
	\caption{Qualitative results of ablation studies.}
	\label{abla1}
\end{figure}

\subsection{Ablation Studies}
In our method, there are three critical designs to ensure the effective functioning of the fusion model: dynamic weights $\alpha_{\{\Lambda, \overline{\Lambda}\}}$ in degradation-aware reconstruction loss, language-feature alignment loss $\mathcal{L}_{ali}$, and CLIPSeg fine-tuning $\Phi^{'}_c$. We conduct ablation studies to validate their roles, as shown in Fig.~\ref{abla1}. First, simultaneously removing $\alpha_{\{\Lambda, \overline{\Lambda}\}}$ and $\mathcal{L}_{ali}$ leads to the worst fusion quality. Second, when only $\mathcal{L}_{ali}$ is removed, the visual performance is slightly improved. Third, removing $\alpha_{\{\Lambda, \overline{\Lambda}\}}$ significantly weakens the removal of flare. These results indicate that $\mathcal{L}_{ali}$ enhances our RFC's generalization across scenarios, while $\alpha_{\{\Lambda, \overline{\Lambda}\}}$ is crucial for flare removal, aligning with our design goals. Besides, if CLIPSeg is not fine-tuned and the original localization output is directly used, flare removal will be insufficient due to inaccurate localization. The quantitative results in Table~\ref{ab-tab} validate the above conclusion, as our RFC achieves the best scores when all these designs are present.

\begin{figure}[t]
	\centering
	\includegraphics[width=0.5\textwidth]{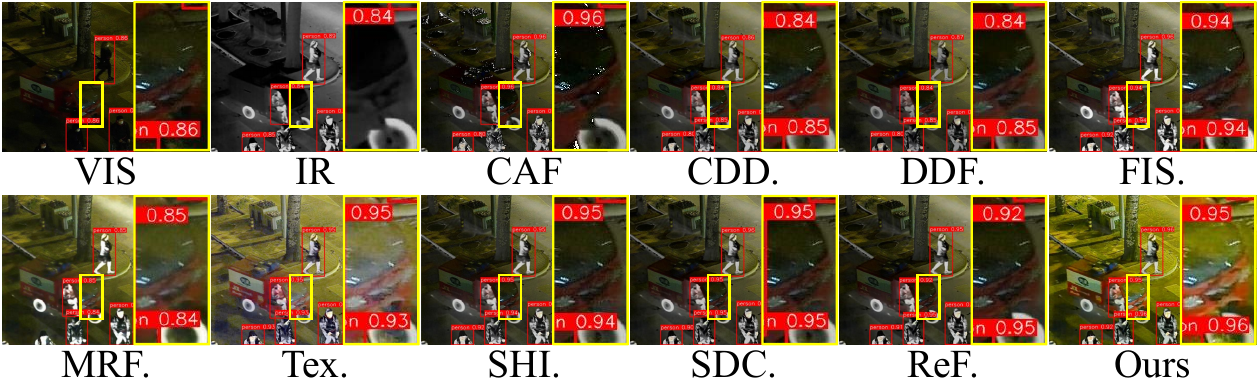}
	\vspace{-0.7cm}
	\caption{Qualitative object detection verification.}
	\label{llvip-de}
\end{figure}

\begin{figure}[t]
	\centering
	\includegraphics[width=0.5\textwidth]{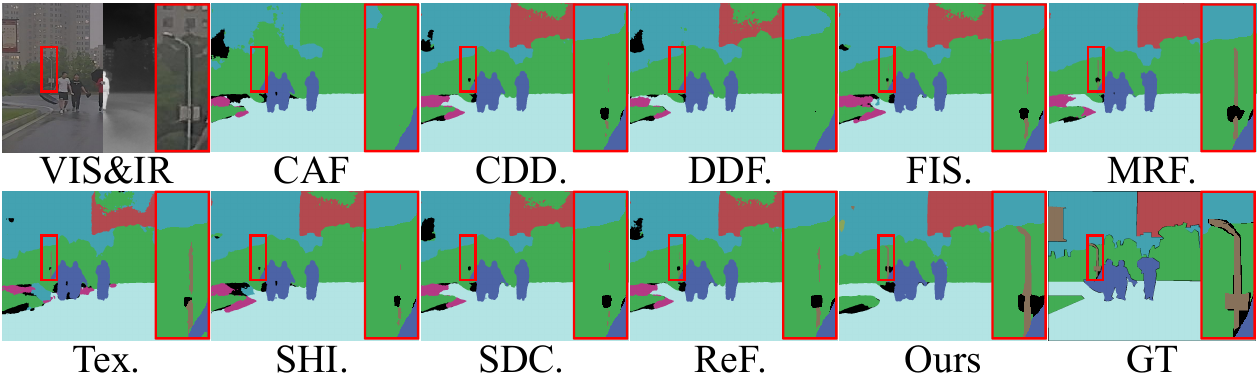}
	\vspace{-0.7cm}
	\caption{Qualitative semantic segmentation verification.}
	\label{fmb-de}
\end{figure}

\subsection{Semantic Verification on High-level Tasks}
$\textbf{Object Detection}$. We implement object detection on the LLVIP dataset with YOLO-v5, in which the detector is retrained on the results of these fusion methods and the source images. As shown in Fig.~\ref{llvip-de} and Table~\ref{de-ta}, our RFC surpasses all competitors in detection accuracy, demonstrating its ability to improve high-level semantic tasks. Notably, detection performance based on the fused images outperforms those based on the source images, highlighting the value of the image fusion technology.

\noindent $\textbf{Semantic Segmentation}$. We retrain SegFormer~\cite{xie2021segformer} on the FMB dataset and apply it to the fused results of each method. As shown in Fig.~\ref{fmb-de} and Table~\ref{fmb-table}, our RFC achieves superior segmentation performance across multiple categories, with results second only to MRFS. The reason is that MRFS couples intermediate features from segmentation and fusion networks, allowing it to retain more semantic information in the fused image.

\begin{table}[t]
    \renewcommand \arraystretch{1.2}
	\caption{Quantitative results of ablation studies (the mean of FMB, LLVIP, M3FD, MFNet datasets).} \vspace{-0.3cm}  \label{ab-tab}
   \resizebox{\columnwidth}{!}
{
	\begin{tabular}{cccc|cccccc}
		\Xhline{1.3pt}
		\rule{0pt}{2.8ex}
		&$\alpha_{\{\Lambda, \overline{\Lambda}\}}$ &  $\mathcal{L}_{ali}$ &$\Phi^{'}_c$ & SSIM & SD & MI & VIF & SCD & Qabf
		 \\             \Xhline{1.3pt}
		\rule{0pt}{2.8ex}
		\MakeUppercase{\romannumeral1} &
		\ding{56} & \ding{56} & \ding{52} & { 0.322} & \cellcolor{blue!10} \underline{44.751} & 2.785                        & { 0.481} & \cellcolor{pink!40} \textbf{ 1.281} & 0.460                  \\
		\MakeUppercase{\romannumeral2}&\ding{56} & \ding{52} & \ding{52}  & 0.319                        & { 44.525} & 2.733                        & 0.473                        & 1.236                        & \cellcolor{blue!10} \underline{0.466}  \\
		\MakeUppercase{\romannumeral3}&\ding{52} & \ding{56} & \ding{52}  & \cellcolor{blue!10} \underline{0.324} & 43.924                        & { 2.815} & \cellcolor{blue!10} \underline{0.484} & \cellcolor{blue!10} \underline{1.242} & { 0.465}  \\
		\MakeUppercase{\romannumeral4}&\ding{52} & \ding{52} & \ding{56}          & 0.321                        & 44.277                        & \cellcolor{blue!10} \underline{2.817} & 0.479                        & 1.226                        & 0.460  \\
		Full-Model&\ding{52} & \ding{52} & \ding{52}  & \cellcolor{pink!40} \textbf{ 0.325} & \cellcolor{pink!40} \textbf{ 45.113} & \cellcolor{pink!40} \textbf{ 2.867} & \cellcolor{pink!40} \textbf{ 0.485} & { 1.242} & \cellcolor{pink!40} \textbf{ 0.469}  \\ \Xhline{1.3pt}
	\end{tabular}
}
\end{table}

\begin{table}[t]
    \renewcommand \arraystretch{1.1}
	\caption{Quantitative object detection verification.}
	\vspace{-0.3cm}
	\label{de-ta}
	\resizebox{\columnwidth}{!}
	{
		\begin{tabular}{c|cccc|cc}
			\Xhline{1.3pt}
			\rule{0pt}{2.8ex}
			\textbf{LLVIP}     & Precision & Recall  & mAP@0.6 & mAP@0.85 & mAP@(0.5:0.95) \\ 		\Xhline{1.15pt}
			\rule{0pt}{2.8ex}
			VIS       & 79.0                          & 62.5                        & 65.4                                                     & 50.6                                                      & 53.5                                                            \\
			IR        & 90.2                          & 78.1                        & 71.2                                                     & 51.0                                                      & 54.7                                                            \\
			DDFM      & 95.6                          & 85.8                        & {85.4}                              & { 61.7}                               & { 71.1}                                     \\
			SHIP      & 93.0                          & 87.1                        & 72.2                                                     & 55.8                                                      & 61.5                                                            \\
			CDDFuse   & 93.9                          & 89.0                        & 80.6                                                     & 52.8                                                      & 65.5                                                            \\
			\cellcolor{blue!10} {MRFS} & \cellcolor{blue!10}{94.1}                        & \cellcolor{blue!10}{82.8}                        & \cellcolor{blue!10}{ 83.3}                              &\cellcolor{blue!10} { 60.7}                               & \cellcolor{blue!10} \underline{74.4}                                     \\
			CAF       & 94.7                          & 86.2                        & 79.2                                                     & 57.7                                                      & 65.9                                                            \\
			ReFusion  & { 96.0}   &{ 91.0} & 77.8                                                     & 55.0                                                      & 60.7                                                            \\
			FISCNet   & 93.6                          & 89.0                        & 72.2                                                     & 52.4                                                      & 56.8                                                            \\
			SDCFusion & 93.2                          & 88.6                        & 72.2                                                     & 53.5                                                      & 56.4                                                            \\
			Text-IF   & {96.0}   &{90.9} & 79.5                                                     & 58.3                                                      & 64.8                                                            \\
			\cellcolor{pink!40}{Ours}      & \cellcolor{pink!40} { 96.7}   & \cellcolor{pink!40}{ 90.2} & \cellcolor{pink!40}{87.5}                              & \cellcolor{pink!40}{61.5}                               & \cellcolor{pink!40} \textbf{ 75.3}      \\ \Xhline{1.3pt}
		\end{tabular}
	}
\end{table}

\begin{table}[]
    \renewcommand \arraystretch{1.2}
	\caption{Quantitative semantic segmentation verification.}
	\vspace{-0.3cm} \label{fmb-table}
	\resizebox{\columnwidth}{!}
	{
		\begin{tabular}{c|cccccccccc}
			\Xhline{1.3pt}
			\rule{0pt}{2.8ex}
			\textbf{FMB}            & DDF.    & SHI.    & CDD.   & \cellcolor{pink!40}{MRF.} & CAF    & ReF.  & FIS.                        & SDC. & Tex.                       & \cellcolor{blue!10}{Ours}                         \\ 			\Xhline{1.15pt}
			\rule{0pt}{2.8ex}
			Vegetation & 42.12                        & 48.67                        & 49.28   & \cellcolor{pink!40}{73.16} & 40.83 & 50.32                        & { 50.89} & 49.76     & 43.64   & \cellcolor{blue!10} { 75.18} \\
			Building   & 56.81                        & 62.01                        & 60.18   & \cellcolor{pink!40}{78.7}  & 55.42 & { 62.4}  & 60.73                        & 59.27     & 53.24   & \cellcolor{blue!10}{ 81.38} \\
			Person     &{50.25} & 48.67                        & 42.69   & \cellcolor{pink!40}{ 52.45} & 41.03 & 42.77                        & 31.7                         & 33.02     & 34.37   & \cellcolor{blue!10}{ 49.86} \\
			Car        & 70.63                        & { 75.07} & 72.61   & \cellcolor{pink!40}{ 77.66} & 70.21 & 74.66                        & 73.91                        & 72.59     & 70.71   & \cellcolor{blue!10} {77.59} \\
			Sky        & 54.39                        & 67.34                        & 69.28   & \cellcolor{pink!40}{ 90.5}  & 50.3  & 69.94                        & { 72.3}  & 71.13     & 58.19   & \cellcolor{blue!10}{89.6}  \\ \Xhline{0.8pt} \rule{0pt}{2.4ex}
			mIoU       & 52.15                        & 56.37                        & 56.61   & \cellcolor{pink!40} \textbf{ 67.54} & 50.64 & { 57.79} & 57.24                        & 57.65     & 51.79   & \cellcolor{blue!10} \underline{67.01} \\ \Xhline{1.3pt}
		\end{tabular}
	}
\end{table}

\section{Conclusion}
This study proposes a robust fusion controller, termed RFC. It can achieve degradation-aware image fusion with fine-grained language instructions, improving the fusion model's robustness in harsh environments with spatial-varying composite degradations. RFC parses language instructions into functional and spatial conditions, and couples them to obtain the composite control priori. With the continuous modulation of this priori on the fusion process, combined with the guidance of the language-feature alignment loss, RFC can ultimately eliminate composite degradations according to language instructions. Extensive experiments demonstrate RFC's superiority in both perceptual performance and semantic quality.

{
    \small
    \clearpage
	\bibliographystyle{ieeenat_fullname}
	\bibliography{main}
}

\end{document}